# Design of Spectrum Sensing Policy for Multi-user Multi-band Cognitive Radio Network

Jan Oksanen, Jarmo Lundén and Visa Koivunen
SMARAD CoE, Department of Signal Processing and Acoustics
School of Electrical Engineering, Aalto University, Finland
Email: jan.oksanen@aalto.fi, jarmo.lunden@aalto.fi, Visa.Koivunen@hut.fi

*Abstract*—Finding an optimal sensing policy for a particular access policy and sensing scheme is a laborious combinatorial problem that requires the system model parameters to be known. In practise the parameters or the model itself may not be completely known making reinforcement learning methods appealing. In this paper a non-parametric reinforcement learning-based method is developed for sensing and accessing multi-band radio spectrum in multi-user cognitive radio networks. A suboptimal sensing policy search algorithm is proposed for a particular multi-user multi-band access policy and the randomized Chair-Varshney rule. The randomized Chair-Varshney rule is used to reduce the probability of false alarms under a constraint on the probability of detection that protects the primary user. The simulation results show that the proposed method achieves a sum profit (e.g. data rate) close to the optimal sensing policy while achieving the desired probability of detection.

*Index Terms*—Cognitive radio, sensing policy, reinforcement learning

## I. Introduction

Spectrum sensing is a key task in cognitive radio (CR). Spectrum sensing is needed for identifying the state of the licensed spectrum (whether it is idle for secondary use or not) as well as for modeling and managing interference. In order to guarantee reliable sensing under demanding fading conditions and to divide the sensing work cooperation is needed among the secondary users (SUs).

In cooperative spectrum sensing the SUs send their local binary decisions (or local test statistics) to a fusion center (FC), that makes the global decision using a fusion rule. The reliability of the global decision is measured by the probability of detection and the probability of false alarm. The probability of detection tells how well the primary users (PUs) can be protected from interference, i.e., situations where the SUs falsely identify the spectrum to be idle and try to access the spectrum simultaneously with the PU. The probability of false alarm on the other hand reflects on how well idle spectrum can be detected. In order to protect the PU while utilizing most of the available spectrum holes, we would like to minimize the probability of false alarm under a constraint on the detection probability.

Depending on the local sensing channel conditions the detection probability at the FC may be either too low so that the interference constraint is violated, or too high so that available spectrum opportunities are being overlooked more than necessary. Assuming that the FC knows or is able to learn the SUs' local sensing performance indices it is possible to use the Chair-Varshney (CV) rule [1] to balance between the global false alarm probability and detection probability. Commonly it is assumed that the values are equal in each detector and these values are selected by the designer. However, there are versions of the CV-rule where the values are estimated [2].

In CR-networks the multi-user multi-band secondary spectrum use is governed by two policies. A sensing policy $\pi_s$ that tells which frequency bands to sense and by which SUs, and an access policy $\pi_a$ which tells when, how and by whom the idle frequencies can be accessed [3]. In this paper we focus only on who gets to access and when. In [4] a decentralized cognitive MAC protocol based on the theory of partially observable Markov decision processes (POMDPs) was proposed, where the SUs independently search for available spectrum holes. In [5] the separation principle that decouples the design of the optimal sensing policy from the design of the access policy and the sensor operating point was established. However, these works do not consider the effect of the distribution of the sensing resources on the sensing policy. In [6] we proposed a reinforcement learning-based sensing policy with a sensing assignment optimization for a fixed fusion rule. In [7] a reinforcement learning based multiagent spectrum sensing policy is proposed that balances between two mutually exclusive objectives: minimizing the false alarm rate under a detection probability constraint and the number of frequency bands sensed simultaneously.

In this paper a reinforcement learning-based multi-user multi-band spectrum sensing and access method is proposed. A heuristic sensing policy search algorithm is proposed for an access policy that allocates SUs to idle bands so that a function balancing between sum data rate and fairness is maximized. For the cooperative sensing the randomized CV-rule is employed. For estimating the local performance indices needed by the CV-rule the proposed reinforcement learning method exploits spatial diversity that makes the global decisions reliable. The contributions of this paper are as follows:

- The randomized CV-rule is used to reduce the number of false alarm under a constraint on the detection probability.
- A heuristic sensing policy search is proposed for the randomized CV-rule and a multi-user multi-band access policy balancing between data rate maximization and user fairness.
- A reinforcement learning method that learns the needed

local detection performance indices, data rates and the band availability probabilities.

The rest of this paper is organized as follows. In section II the system model for the SU and PU networks is described. In section III the randomized CV-rule is presented. In section IV a simple access policy is developed that facilitates balancing between the SU network sum data rate maximization and fairness. In section V the optimal sensing policy problem is formulated when all system model parameters are known. In section VI a reinforcement learning-based sensing and access method is proposed along with a heuristic sensing policy search algorithm. Section VII shows simulation examples of the proposed method against a genie aided optimal sensing policy. The paper is concluded in section VIII.

## II. SYSTEM MODEL

In this paper the spectrum of interest is assumed to be divided into $K$ bands that are allocated for different independent PUs. Each band is assumed to be idle with an unknown probability $P_k$, where $k$ is the band index.

In the SU network distributed spectrum sensing governed by an FC is assumed as shown in Fig. 1. The FC may be part of the network infrastructure (such as a base station) or one of the SUs may act as the FC. There are $N$ spatially disperse SUs equipped with Neyman-Pearson detectors with known false alarm rates cooperatively sensing the $K$ frequency bands. The SUs operate in a time slotted fashion as shown in Fig. 2. In the sensing minislot each SU senses up to $K_i$ bands assigned by a sensing policy $\pi_s$ and makes a local binary decision about the availability of the band(s). In the beginning of the communication minislot the SUs send their local binary decisions to an FC via a common control channel. During the communication minislot the FC makes the global decision using the randomized CV-rule and grants the SUs access to the possibly found idle spectrum using an access policy $\pi_a$. At the end of the communication slot the FC selects the next set of bands to be sensed and the corresponding sensing assignment using the sensing policy $\pi_s$ and then signals this information along with the spectrum access information to the SUs. It is assumed that the SUs have always data to send and when an SU $i$ is granted access to an idle band $k$ the it achieves a data rate $R_{i,k}$. When the SU gets spectrum access but the band is occupied by the PU a collision will take place and the achieved data rate becomes 0. In the next communication slot the SUs who were granted access will feed back their achieved data rates (or estimated rates) to the FC.

## III. SENSING SCHEME: RANDOMIZED CHAIR-VARSHNEY FUSION RULE

In order to simplify the notation momentarily, assume that the CR-network is sensing only one frequency band. To minimize the probability of false alarm under the constraint on the detection probability, the CV-rule [1] may be used. The CV-rule weights the local decisions according to the local detection probabilities $\beta_i$ and false alarm probabilities $\alpha_i$. Typically these probabilities are not available, but for now assume that they are known. A simple reinforcement learning method for estimating these probabilities is presented later in section VI.

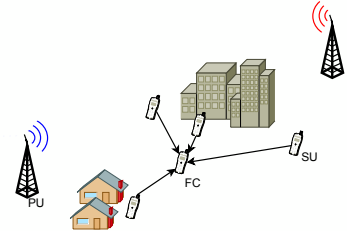

Fig. 1. A CR setting. The SUs are collaboratively sensing whether the PUs are active or not. After sensing the spectrum the SUs send their local binary decisions to a fusion center (FC) that makes a global decision about the state of the spectrum and grants access for one of the users if the band is found to be idle. Cooperative spectrum sensing provides spatial diversity that mitigates the effects of fading caused by large objects and fast fading caused by multi-path propagation and mobility.

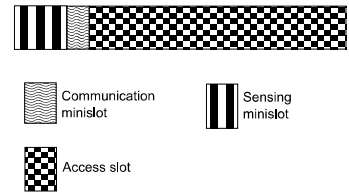

Fig. 2. Slotted operation of the CR-network. In the sensing minislot the SUs sense the spectrum according to the sensing policy $\pi_s$. In the communication slot they then send their local binary decisions through a common control channel to the FC which makes a global decision about the availability of the spectrum using a fusion rule (in this paper the CV-rule). Then the FC grants access to the spectrum for the SUs according to an access policy $\pi_a$ and assigns the SUs for sensing for the next sensing slot. The achieved data rates are communicated to the FC in the next communication slot.

The SUs send their local binary decisions $u_i$ to an FC that makes a global decision about the presence of a primary signal. Let $H_0$ denote the null hypothesis that the PU is not transmitting and $H_1$ denote the alternative hypothesis that the PU is transmitting. Given the vector $\mathbf{u}$ of local decisions the FC makes the global decision by computing a likelihood ratio test (LRT) as [1]

$$T(\mathbf{u}) = \frac{\Pr\{u_1, u_2, ..., u_N \mid H_1\}}{\Pr\{u_1, u_2, ..., u_N \mid H_0\}} \underset{u_{i^*}=0}{\overset{u_{i^*}=1}{\gtrless}} \eta, \quad (1)$$

where $i^*$ is the index of the SU acting as the FC and $\eta$ the decision threshold. Assuming conditional independence among $u_i$'s the LRT can be expressed as [1]

$$T(\mathbf{u}) = \sum_{i=1}^{N} \left[ u_i \log \frac{\beta_i(1-\alpha_i)}{\alpha_i(1-\beta_i)} + \log \frac{1-\beta_i}{1-\alpha_i} \right] \underset{u_{i^*}=0}{\overset{u_{i^*}=1}{\gtrless}} \eta. \quad (2)$$

By tuning the threshold $\eta$ one can control the global probability of false alarm and the global probability of detection. Assuming that $\beta_i > \alpha_i$ the threshold can be chosen such that the probability of detection at the FC is above the desired

value $\beta$, as
$$\check{\eta} = \sup\left\{\eta : \Pr\{T(\mathbf{u}) > \eta | H_1\} \geq \beta\right\}. \quad (3)$$

Then the probability of false alarm becomes $\check{\alpha}_{i^*}(\mathbf{u}) = \Pr\{T(\mathbf{u}) > \check{\eta} | H_0\}$.

According to the Neyman-Pearson lemma the probability of false alarm at the FC is minimized if the detection probability constraint in (3) is met exactly. Since we are dealing with discrete random variables, it is not always possible to achieve the desired detection probability exactly. To meet the detection probability constraint exactly, randomization can be used, which however does not anymore guarantee optimality in the Neyman-Pearson sense [8]. However, as we will show it still allows us to lower the false alarm probability under the detection probability constraint compared to the non-randomized CV-rule given in (3). The randomized decision rule is given as

$$\Pr\{u_{i^*} = 1\} = \begin{cases} 1, & \text{if } T(\mathbf{u}) > \hat{\eta}, \\ \rho, & \text{if } T(\mathbf{u}) = \hat{\eta}, \\ 0, & \text{if } T(\mathbf{u}) < \hat{\eta}, \end{cases} \quad (4)$$

where

$$\rho = \frac{\beta - \Pr\{T(\mathbf{u}) > \hat{\eta} | H_1\}}{\Pr\{T(\mathbf{u}) = \hat{\eta} | H_1\}} \quad (5)$$

and $\hat{\eta} = \inf\{\eta : \Pr\{T(\mathbf{u}) > \eta | H_1\} \leq \beta\}$. The idea of the randomized CV-rule is to select the decision threshold as the smallest $\eta$ achieving a detection probability below the desired level $\beta$ and then use randomization to achieve $\beta$ exactly. The probability of false alarm at the FC using randomized CV-rule becomes then

$$\hat{\alpha}_{i^*}(\mathbf{u}) = \Pr\{T(\mathbf{u}) > \hat{\eta} | H_0\} + \rho \Pr\{T(\mathbf{u}) = \hat{\eta} | H_0\}. \quad (6)$$

It is easy to show that the false alarm probability at the FC using the randomized CV-rule given in (4)-(5) is always less or equal to the false alarm probability of the non-randomized CV-rule given in (2)-(3):

$$\check{\alpha}_{i^*}(\mathbf{u}) = \Pr\{T(\mathbf{u}) > \check{\eta} | H_0\}$$
$$= \Pr\{T(\mathbf{u}) > \hat{\eta} | H_0\} + \Pr\{T(\mathbf{u}) = \hat{\eta} | H_0\} \geq \hat{\alpha}_{i^*}(\mathbf{u}).$$

## IV. ACCESS POLICY

When idle spectrum is discovered the FC must decide whether to access the spectrum or not. Denote the probability of accessing band $k$ when the sensing outcome at the FC has been $u_{i^*,k}$ as $a_k(u_{i^*,k})$. Then the probability of collision with the PU under $H_1$ is $a_k(u_{i^*,k})(1 - \beta_{i^*,k})$ where $\beta_{i^*,k}$ is the probability of detection at the FC at band $k$. When the spectrum regulator requires that the probability of collision should be no more than $\omega$, then the optimal operating point in terms of throughput is $\beta_{i^*,k} = 1 - \omega$ and $a_k(u_{i^*,k}) = u_{i^*,k}$ (see [5] theorem 2). Note that making first the decision about the availability of the spectrum and then randomizing the access decision can always be translated into randomizing first the sensing decision and then trusting it in the access. For instance in this paper the randomized CV-rule is used.

When idle spectrum is discovered the FC decides which SUs get access to the spectrum. Assume that the FC knows or is able to learn the expected data rates $R_{i,k}$ when SU $i$ accesses band $k$ and band $k$ is idle. Furthermore, assume that the SUs have always data to send. Denote the weighted average data rate that SU $i$ has obtained so far as $J_i$ (counting also the cases where SU $i$ got no access). Assume that the FC has sensed a set of subbands $\mathcal{L}$ to be idle. One possible way to allocate the SUs to the bands in $\mathcal{L}$ is to maximize a function balancing between sum data rate and fairness as

$$y_{i,k} = \arg\max_{y_{i,k}} \sum_{i=1}^{N} \sum_{k \in \mathcal{L}} y_{i,k} \frac{R_{i,k}^{\theta}}{J_i^{\nu}} \quad (7)$$

$$\text{s.t.} \sum_{i=1}^{N} y_{i,k} = 1, \quad \text{s.t.} \sum_{k \in \mathcal{L}} y_{i,k} = 1. \quad (8)$$

In (7) and (8) $y_{i,k} = 1$ if SU $i$ is granted access to band $k$ and $y_{i,k} = 0$ otherwise. The constraints in (8) require that each SU can only get access to one band at a time and that each band may be accessed only by one SU at a time. Parameters $\nu$ and $\theta$ control the balance between sum data rate maximization and fairness, i.e., granting the access to the best SUs with the currently best channel conditions and granting the access to all SUs equally fairly. When $\theta = 1$ and $\nu = 0$ the access policy maximizes the expected sum data rate in the SU network, and when $\theta = 0$ and $\nu = 1$ the policy distributes the data rate equally among the SUs. This is a linear assignment problem (or maximum weight matching problem) that may be solved in polynomial time using the Hungarian algorithm [9].

More sophisticated access policies tailored for a particular type of data and network designs could be employed, but in order to keep the presentation simple in this paper we adopt the access policy given in (7)-(8).

## V. SENSING POLICY

### A. Optimal sensing policy

In this paper the optimal spectrum sensing policy $\pi_s^*$ is defined as the policy that selects the optimal frequency bands to be sensed and the optimal SUs to sense them such that the expected immediate profit from the employed access policy is maximized. The profit from the access policy may be associated with the obtained sum data rate or it may be a function of the data rate and a measure of fairness (for example as the policy defined in (7)-(8)). Without loss of generality in the following the profit from the access policy is denoted as the achieved data rate. The optimal sensing policy in this paper is a genie aided policy that knows all the model parameters and that has no computation time and memory limitations. Furthermore, the randomized CV-rule is assumed to be used with $\beta_{i^*,k} = 1 - \omega$ and consequently $a_k(u_{i^*,k}) = u_{i^*,k}$.

Denote the set of sensed subbands that the FC has selected following a sensing policy $\pi_s$ as $\mathcal{K}(\pi_s)$. Once the FC has discovered a subset of idle bands $\mathcal{L} \in \mathcal{K}(\pi_s)$ it grants access to one of the SUs at each idle subband using an access policy $\pi_a$. Denote the expected sum data rate after applying access policy $\pi_a$ to $\mathcal{L}$ as $\mathrm{E}[R_\mathcal{L}(\pi_a)]$. Then the expected sum data rate from the bands $\mathcal{K}(\pi_s)$ is

$$R_{\mathcal{K}(\pi_s)}(\pi_a) = \sum_{\mathcal{L} \in \mathcal{K}(\pi_s)} \left( \mathrm{E}[R_\mathcal{L}(\pi_a)] \cdot \prod_{k \in \mathcal{L}} \psi_k \cdot \prod_{\substack{j \in \mathcal{K} \\ j \notin \mathcal{L}}} (1-\psi_j) \right) \quad (9)$$

where the summation is over all possible subsets $\mathcal{L}$ of $\mathcal{K}(\pi_s)$. $\psi_k$ is the probability that the FC detects band $k$ to be idle and is given by

$$\psi_k = \left(1 - \alpha_{i^*,k}(\pi_s)\right) P_k + \omega(1 - P_k), \quad (10)$$

where $P_k$ is the a priori probability that band $k$ is idle. $\mathrm{E}[R_\mathcal{L}(\pi_a)]$ is computed using the Bayes' formula as

$$\mathrm{E}[R_\mathcal{L}(\pi_a)] = \sum_{k \in \mathcal{L}} \frac{R_k(\pi_a)(1 - \alpha_{i^*,k}(\pi_s)) P_k}{(1 - \alpha_{i^*,k}(\pi_s)) P_k + \omega(1 - P_k)}, \quad (11)$$

where $R_k(\pi_a)$ is the achievable data rate at band $k \in \mathcal{L}$ when access policy $\pi_a$ is applied to $\mathcal{L}$. The false alarm probabilities are functions of the sensing policy since they depend on which SUs are assigned to sense which band. Requiring that each SU can sense only up to $K_i$ bands simultaneously the optimal sensing policy becomes

$$\pi_s^* = \arg\max_{\pi_s} R_{\mathcal{K}(\pi_s)}(\pi_a) \quad (12)$$

$$\text{s.t. } K_i(\pi_s) \leq K_i, \quad (13)$$

where $K_i(\pi_s)$ is the number of bands sensed by SU $i$ using sensing policy $\pi_s$. Note that the probability of detection constraint is included implicitly in the CV-rule and hence is not visible here. Once the design of the sensing scheme and the access policy $\pi_a$ are fixed the optimal sensing policy becomes essentially a function of $\pi_a$ and the sensing scheme. This is intuitive since once it is known how the idle bands would be used and what are the sensing capabilities of individual SUs one can optimize the sensing strategy. It can be seen that computing the optimal sensing policy is a difficult combinatorial problem and hence we focus on a heuristic (possibly suboptimal) search.

## VI. Reinforcement learning-based sensing and access

Since the detection probabilities, PU transmission probabilities and the achievable data rates are generally not known they need to be estimated. In this paper we propose a simple reinforcement learning-based sensing and access method that is non-parametric to the PU activity and achievable data rate models. The proposed method employs action value-learning with one state and $\epsilon$-greedy exploration [10]. With probability $\epsilon$ the method chooses random actions to learn about the values of all possible actions and with probability $1 - \epsilon$ it tries to maximally exploit its current knowledge by choosing the seemingly best actions. The policy has three types of actions: selecting the bands to be sensed, assigning an SU to sense a certain band and granting access to an SU at a certain band. Next we will discuss how the estimated values of the actions are updated and how the actions are selected in the exploration and exploitation phases.

### A. Action value updates

The value update after taking an action $a$ at time instant $t$ is given by exponential smoothing as [10]

$$Q_{a,t+1} = Q_{a,t} + \delta_a \left[ r_{a,t+1} - Q_{a,t} \right], \quad (14)$$

where $\delta_a$ is the step size and $r_{a,t+1}$ is the obtained reward after taking action $a$. In this paper constant a step size $\delta_a$ is used since in practise the sensing policy problem is non-stationary. A constant $\delta_a$ does not guarantee the estimated values to converge to the true values with probability 1 but only in expectation [6]. However, this is an unavoidable concession when tracking non-stationary problems. The $Q$-values that the FC keeps track of are denoted as:

| $Q^{sen}_{i,k,t}$ | Value of assigning SU $i$ to sense band $k$ |
| --- | --- |
| $Q^{acc}_{i,k,t}$ | Value of granting SU $i$ access to band $k$ |
| $Q_{k,t}$ | Value of assuming band $k$ to be idle |

The reward $r^{acc}_{i,k,t+1}$ from allocating SU $i$ to band $k$ at time $t+1$ is

$$r^{acc}_{i,k,t+1} = \begin{cases} R_{i,k,t+1}, & \text{if } y_{i,k} = 1 \text{ and no collision} \\ Q^{acc}_{i,k,t}, & \text{otherwise,} \end{cases} \quad (15)$$

where $R_{i,k,t+1}$ is the achieved instantaneous data rate when SU $i$ accessed band $k$ at time $t+1$. A collision takes place when a missed detection occurs at the FC and one of the SUs tries to access the band simultaneously with the PU. This reward function makes $Q^{acc}_{i,k,t+1}$ to become an estimate of the mean achievable data rate at band $k$ by user $i$.

The reward function for estimating the probabilities of band $k$ being idle is given by

$$r_{k,t+1} = \begin{cases} 0, & \text{if } u_{i^*,k,t+1} = 1 \text{ or collision} \\ 1, & \text{if } u_{i^*,k,t+1} = 0 \text{ and no collision} \\ Q_{k,t}, & \text{Otherwise,} \end{cases} \quad (16)$$

This reward function makes $Q_{k,t+1}$ to become an estimate for the probability of band $k$ being idle.

The reward $r^{sen}_{i,k,t+1}$ for assigning SU $i$ to sense band $k$ at time $t+1$ is given by

$$r^{sen}_{i,k,t+1} = \begin{cases} u_{i,k,t+1}, & \text{if } u_{i^*,k,t+1} = 1 \text{ or collision} \\ Q^{sen}_{i,k,t}, & \text{otherwise,} \end{cases} \quad (17)$$

where $u_{i,k,t+1}$ is the local sensing decision at the $i$th SU at band $k$ at time instant $t+1$ and $u_{i^*,k,t+1}$ is the corresponding decision at the FC. When the local decision agrees with the FC that the PU is present the Q-value of the SU is increased towards 1. When the local decision indicates that the band is idle and the decision at the FC indicates that the band is occupied the value is shifted towards 0. If a missed detection occurs at the FC and one of the SUs tries to access the band simultaneously with the PU a collision will take place making the achieved throughput to be 0. Also in this case the local detection probability estimates are shifted either towards 1 or 0 according to the local decision. Whenever the FC thinks that the primary signal is not present at band $k$ the detection probability estimates are kept unchanged.

*B. Exploration by random sensing and access*

With probability $\epsilon$ the learning method goes in to exploration phase. In the exploration phase random actions are taken, i.e., random sensing and access policies are executed. In practise a more systematic approach can be taken and the actions could be made pseudorandom. However, to keep the presentation compact in this paper we use uniformly distributed random actions. For the access policy this means that SUs are allocated randomly to the found idle bands. In the sensing policy the bands to be sensed and the corresponding sensing assignments are picked randomly with a fixed diversity order $D$, where $D$ is the number of SUs simultaneously sensing a band. Diversity guarantees reliability in the sensing needed to obtain good estimates for $Q_{i,k}^{sen}$ and $Q_k$. The idea in exploration is to avoid using the information obtained from the past observations, which might offset the action value estimates. In order to avoid this happening to the probability estimates $Q_{i,k}^{sen}$ and $Q_k$ equal weight decision fusion, i.e., the m–out–of–n rule is used. It is easy to show that with $P_k = 0.5$ and all SUs assumed to have identical detectors (i.e., $\beta_{i,k} = \beta'$ and $\alpha_{i,k} = \alpha'$) the m–out–of–n rule with $m = \left\lceil \frac{D \log \frac{1-\alpha'}{1-\beta'}}{\log \frac{\beta'(1-\alpha')}{\alpha'(1-\beta')}} \right\rceil$, minimizes the probability of error (sum of the probability of missed detection and false alarm) at the FC [1]. In this paper in the exploration phase the local detection probability estimates $Q_{i,k}^{sen}$ are momentarily replaced by $\beta' = \frac{1+\alpha'}{2}$, i.e., the mean of a uniform random variable between $[\alpha', 1]$. Consequently the $Q_{i,k}^{sen}$- and $Q_k$-values are updated only during the exploration phase.

*C. Exploitation by a heuristic sensing policy search*

With probability $1 - \epsilon$ the proposed learning method goes in to exploitation phase. In the exploitation phase the FC tries to maximally use the learned knowledge from the past observations.

As it was seen in section V-A finding the optimal sensing policy is in general a tedious combinatorial problem. In this paper a suboptimal search algorithm (listed in Algorithm 1) is proposed that has low computation and memory requirements. The idea of the proposed policy is to select the set of bands to be sensed by evaluating different $\min\{N, K\}$-size candidate sets and finding a feasible sensing assignment for them by solving multiple consecutive linear assignment problems (step 3). The goodness of a candidate set is evaluated as the sum of all data rate estimates weighted by the estimated probability of the bands being idle and also detected idle (step 4). The considered candidate sets of bands to be sensed are selected as the ones with the largest data rate estimate sum weighted by the sum of one minus the local sensing error probability estimates and the probability estimates of the band being idle

**Algorithm 1**

- STEP 1: Initialize the number of sensed bands as $V = |\mathcal{K}| = \min\{N, K\}$.
- STEP 2: Select the $V$ bands to be sensed as the bands with the largest values of $Q_k \sum_{i=1}^{N} \frac{Q_{i,k}^{acc^\theta}}{J_i^\nu} \sum_{i=1}^{N}(Q_{i,k}^{sen} - \alpha_{i,k})$.
- STEP 3: Find a sensing assignment $\mathbf{X}_\mathcal{K}$ by iteratively assigning one SU to each selected band using the Hungarian algorithm [9] with costs: $-(Q_{i,k}^{sen} - \alpha_{i,k})Q_k \sum_{i=1}^{N} \frac{Q_{i,k}^{acc^\theta}}{J_i^\nu}$ until all SUs are assigned.
- STEP 4: Calculate the probabilities of false alarm $\alpha_{i^*,k}$ using the randomized CV-rule with the constraint $\beta_{i^*,k} = 1 - \omega$ using the sensing assignment obtained in step 3. Then calculate: $E_\mathcal{K} = \sum_{k \in \mathcal{K}} Q_k(1 - \alpha_{i^*,k}) \sum_{i=1}^{N} \frac{Q_{i,k}^{acc^\theta}}{J_i^\nu}$.
- STEP 5: Set $V \leftarrow V - 1$ and repeat steps 2 to 5 until $V = 0$.
- STEP 6: Return the set of bands to be sensed as $\mathcal{K}^* = \arg\max E_\mathcal{K}$ and the corresponding sensing assignment $\mathbf{X}_{\mathcal{K}^*}$.

(step 2). At each round the size of the candidate set is reduced by one until the size becomes zero. The reduction of the candidate set size facilitates the possibility of sensing less bands with higher diversity and consequently with lower false alarm probability. Finally (step 6) the set of bands and the corresponding sensing assignment is selected as the one with the largest value evaluated in step 4.

When idle bands are discovered they are allocated to the SUs according to the access policy given in (7)-(8) using the Hungarian method.

## VII. SIMULATION EXAMPLE

We simulated the proposed reinforcement-based sensing and access method for a small network with $N = 4$ and $K = 3$. The PU activity at each band is modeled as an independent Bernoulli process with probabilities of being idle $P_1 = 0.41$, $P_2 = 0.17$ and $P_3 = 0.50$. The mean local detection probabilities at each SU for each band are given in table I. The local detectors are assumed to be Neyman-Pearson detectors with known false alarm probabilities $\alpha_{i,k} = 0.01$. The achievable mean data rates at each band for each SU when the bands are idle are given table II. The simulation was run for 3 different exploration probabilities $\epsilon = 0.1$, $\epsilon = 0.05$ and $\epsilon = 0.03$. The step sizes for the two probability estimates are $\delta = 0.01$ and for the data rate estimates $\delta = 0.1$. The diversity order in the exploration phase is $D = 2$, making the fusion rule to become the simple OR-rule. In the exploitation phase algorithm 1 is used to find a set of bands to be sensed and the corresponding sensing assignment for the randomized CV-rule. In the exploitation phase the access policy defined in (7)-(8) is used with parameters $\theta = 1$ and $\nu = 0$, i.e., the access policy is maximizing the sum data rate in the SU network. The collision probability is constrained to be $\omega = 0.1$.

Figure 3 shows the achieved sum data rate of the proposed method as a function of time compared to an optimal genie aided sensing policy. Curves are shown to the three different values of $\epsilon$. The genie aided sensing policy is assumed to know the system model parameters and is able to select the best bands to be sensed and the corresponding sensing assignments that maximize the sum data rate. It can be seen that the proposed method achieves in this case approximately 90% of the optimal sum data rate. After the initial transition the method is momentarily caught in a local maxima, but is eventually able to find its way out to a better solution.

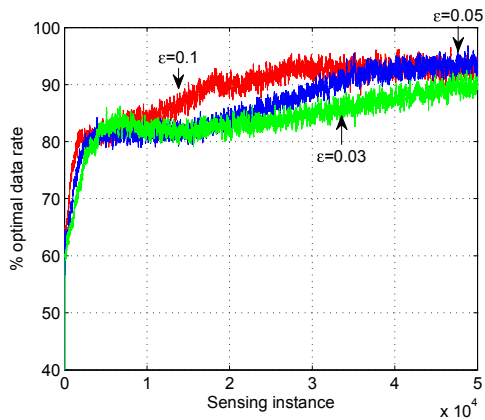

Fig. 3. The achieved sum data rate in the SU network relative to an optimal genie aided policy with known parameters. After the initial transition the method is caught in a local maxima, but eventually reaches close to 90% of the optimal sum data rate. The transition from the local maxima to a better solution is faster for large $\epsilon$. However, in steady state small values of $\epsilon$ produce higher data rate.

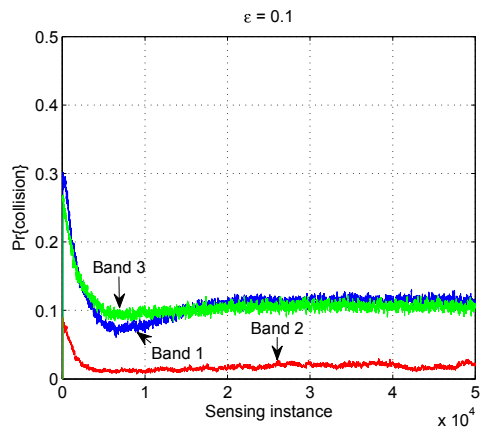

Fig. 4. The probability of collision under $H_1$ at the 3 bands as a function of time when $\epsilon = 0.1$. The collision probability a band 1 and 3 approach to the desired value $\omega = 0.1$. Since in steady state band 2 is basically sensed only during the exploration phases (due to its lower availability and data rates), the collision probability at band 2 is small.

Figure 4 shows the collision probability under $H_1$ at the three bands as a function of time for $\epsilon = 0.1$. After convergence the sensing policy chooses to sense only bands 1 and 3 in the exploitation phase. It can be seen that at these two bands the collision probability approaches the desired value $\omega = 0.1$. For band 2 the collision probability is clearly below $0.1$ since it is practically only sensed during exploration phases. Due to the exploration phase with OR-rule the collision probabilities at bands 1 and 3 are slightly biased from the desired value $0.1$. However, for small $\epsilon$ this bias is expected to be small. For large $\epsilon$ if the bias is unacceptably large the SUs may be denied to access during the exploration phases.

TABLE I
THE MEAN LOCAL $\beta_{i,k}$'S AT EACH BAND FOR EACH SU.

| $\beta_{i,k}$ | Band 1 | Band 2 | Band 3 |
|---|---|---|---|
| SU 1 | 0.53 | 0.93 | 0.14 |
| SU 2 | 0.16 | 0.70 | 0.78 |
| SU 3 | 0.18 | 0.42 | 0.50 |
| SU 4 | 0.66 | 0.83 | 0.52 |

TABLE II
MEAN DATA RATES AT EACH BAND FOR EACH SU.

| $R_{i,k}$ | Band 1 | Band 2 | Band 3 |
|---|---|---|---|
| SU 1 | 67.9 | 75.0 | 45.5 |
| SU 2 | 4.0 | 13.9 | 75.0 |
| SU 3 | 60.0 | 3.9 | 51.1 |
| SU 4 | 36.8 | 23.7 | 99.2 |

## VIII. CONCLUSIONS

In this paper a spectrum sensing and access method based on reinforcement learning has been proposed for multi-user multi-band CR. The method uses the randomized CV-rule in order to reduce the number of false alarms under a detection probability constraint. The optimal sensing policy for a particular access policy and sensing scheme can be found via an exhaustive search. However, such a search is not computationally feasible in practice. In this paper a simple and fast suboptimal sensing policy search algorithm has been proposed for the CV-rule and an access policy that allows for balancing between data rate maximization and fairness. The simulation results have illustrated the performance of the proposed method to be close to a genie aided policy.